\PassOptionsToPackage{colorlinks=true, linkcolor=blue, citecolor=blue, urlcolor=blue}{hyperref}

\documentclass[letterpaper, 10 pt, journal, twoside]{IEEEtran}

\usepackage{amsfonts}
\usepackage{amsmath}
\usepackage{amssymb}
\usepackage{bm}
\usepackage[table]{xcolor}
\usepackage{booktabs}
\usepackage{cite}
\usepackage{orcidlink}
\usepackage{cleveref}
\usepackage{enumitem}
\usepackage{multicol}
\usepackage{graphicx}
\usepackage{siunitx}
\usepackage[percent]{overpic}

\Crefname{figure}{Fig.}{Figs.}
\Crefname{equation}{Eq.}{Eqs.}

\newcommand{\Tr}{^{\top}}
\newcommand{\paral}{{\mkern3mu\vphantom{\perp}\vrule depth 0pt\mkern3mu\vrule depth 0pt\mkern3mu}}

\begin{document}
\title{Contact-Implicit Whole-Body Trajectory Optimization for Dynamic Humanoid Locomotion on Parameterizable Uneven Terrain}

\author{Roberto~Mauceri$^{\, \orcidlink{0009-0006-3291-9232}}$,
       Stefano~Dafarra$^{\, \orcidlink{0000-0002-1575-5188}}$,
       Giulio~Romualdi$^{\, \orcidlink{0000-0002-6461-3231}}$,
       Marco~Gabiccini$^{\, \orcidlink{0000-0001-5670-892X}}$,
       and~Daniele~Pucci$^{\, \orcidlink{0000-0002-7600-3203}}$%

\thanks{Manuscript received April, 21, 2025; Revised June, 19, 2025; Accepted September, 5, 2025.}%
\thanks{This paper was recommended for publication by Editor A. Kheddar upon evaluation of the Associate Editor and Reviewers’ comments.}%
\thanks{Roberto Mauceri, Stefano Dafarra, Giulio Romualdi and Daniele Pucci are with the Artificial and Mechanical Intelligence, Istituto Italiano di Tecnologia, 16163 Genoa, Italy. Roberto Mauceri and Daniele Pucci are also with the Dept. of Computer Science, University of Manchester, Manchester M13 9PL, U.K.
        {\footnotesize (e-mail: roberto.mauceri@iit.it; stefano.dafarra@iit.it; giulio.romualdi@iit.it; daniele.pucci@iit.it)}.}%
\thanks{Marco Gabiccini is with the Dept. of Civil and Industrial Engineering, University of Pisa, 56126 Pisa, Italy
        {\footnotesize (e-mail: m.gabiccini@ing.unipi.it)}.}%
\thanks{This author-corrected version fixes typos in Eqs. (7)--(8) and includes minor textual clarifications. The implementation and results are unchanged.}%
\thanks{© 2025 IEEE. Personal use of this material is permitted. Permission from IEEE must be obtained for all other uses, in any current or future media, including reprinting/republishing this material for advertising or promotional purposes, creating new collective works, for resale or redistribution to servers or lists, or reuse of any copyrighted component of this work in other works.}%
}

\markboth{%
  \protect\parbox{0.94\textwidth}{\centering
    Author's corrected version of: IEEE Robotics and Automation Letters,
    vol.~10, no.~11, pp.~11674--11681, Nov.~2025.
    \protect\\
    DOI: 10.1109/LRA.2025.3614066
  }%
}{Mauceri \MakeLowercase{\textit{et al.}}:
Dynamic Humanoid Locomotion on Parameterizable Uneven Terrain}

\maketitle

\begin{abstract}
This letter presents a trajectory planner for humanoid robots to perform dynamic motions on complex, parameterizable terrains using a contact-implicit whole-body optimization framework. The planner generates feasible movements without predefined contact sequences, considering the robot's full dynamics and kinematics. To plan bipedal locomotion on steps, ramps, and other terrains, we propose a mathematical model to approximate ground geometry with a smooth surface. The rigid interaction between the robot and the ground is modeled via novel contact constraints, extending the domain of applicability from flat and horizontal surfaces to any arbitrary smooth surface. Effectiveness is demonstrated using nonlinear programming to plan motions on various terrain topologies. We also explore dynamic trajectory generation by embedding high-level requirements in the cost function and the initial guess. Preliminary feasibility tests were conducted over non-coplanar terrains, both in simulation and on the real robot ErgoCub.
\end{abstract}


\section{Introduction}
\label{sec:introduction}
\IEEEPARstart{H}{umanoid} motion planning aims to generate coordinated, physically consistent trajectories that respect system dynamics and contact constraints. While early methods focused on flat ground, real-world applications require handling \emph{uneven terrain}, which demands a more sophisticated approach to selecting feasible contact locations and generating dynamic trajectories. When operating offline, the planner must anticipate variations in terrain geometry and generate feasible motions that maintain stability without real-time feedback by explicitly integrating terrain and contact models.

A successful physics-based strategy for motion planning is \emph{trajectory optimization} (TO), which selects, from the infinite set of physically feasible trajectories, those that globally or locally minimize metrics encoding a desired behavior \cite{kuindersma2016optimization}. TO offers a high level of abstraction and provides powerful tools for the explicit management of contact constraints (e.g., \emph{mixed-integer convex programming} (MICP) \cite{ding2018single} or \emph{linear complementarity problem} \cite{posa2014direct}). It has proven effective in robotics fields that require creating and breaking contacts with the environment, such as manipulation \cite{mordatch2012contact, gabiccini2018computational, aceituno2020global} and legged locomotion \cite{posa2014direct, dai2014whole, deits2014footstep, neunert2018whole, mastalli2020motion, mastalli2020crocoddyl, aceituno2017simultaneous, tonneau2020sl1m, dafarra2022dynamic,jenelten2022tamols, kim2024contact}. Common terrain modeling approaches for legged locomotion rely on planar region extraction from point clouds to simplify foothold planning \cite{bertrand2020detecting, sun2024multi}, while others explore continuous terrain representations \cite{winkler2018gait} or discretized heightmaps \cite{jenelten2022tamols}, enabling richer geometric descriptions.

Traditional TO approaches rely on \emph{predefined contacts}, where contact locations and timings are fixed in advance. In \cite{dai2016planning}, the authors explored robust walking on uneven terrain using \emph{convex programming} (CP), which ensures global optimality. Given predefined footstep locations and timings, their approach regulates the robot centroidal dynamics (CD) while satisfying the contact wrench cone criterion \cite{hirukawa2006universal, qiu2011human}. Footholds are selected based on the method proposed in \cite{deits2014footstep}. The terrain is modeled as stepping stones with flat convex regions where foot placement is permitted. Recent work has advanced these ideas by integrating contact location determination and motion planning into a unified CP framework, addressing challenges of terrain adaptability and dynamic motion for quadrupeds \cite{aceituno2017simultaneous} (\Cref{table:comparison}). To mitigate the combinatorial complexity of MICP, \cite{tonneau2020sl1m} leveraged the sparsity properties of L1-norm minimization and a center-of-mass (CoM) model to reformulate contact planning as a feasibility linear program. Their approach, validated in humanoid simulations, shows promise for online planning on non-flat terrain (\Cref{table:comparison}).

To overcome the limitations of predefined contacts, \emph{contact-implicit} approaches let the planner freely determine them, offering greater flexibility and adaptability \cite{kim2024contact}. A common technique is the use of \emph{complementarity conditions} (CCs), which encode \emph{rigid} contacts and allow smooth transitions between free motion and contact phases \cite{posa2014direct, dai2014whole}. However, CCs are non-convex and introduce nonlinearities, making optimization challenging and often ill-conditioned. Several strategies address these issues: relaxed CCs add a small tolerance in the contact constraints \cite{manchester2019contact}, penalty-based approaches penalize CC violations within the cost function \cite{mordatch2012contact}, and dynamic CCs reformulate the conditions using Baumgarte stabilization \cite{dafarra2022dynamic}.

\emph{Nonlinear programming} (NLP) provides an alternative to MICP when models deviate from convexity. Convex formulations typically rely on simplified or reduced models to preserve convexity, so they cannot capture whole-body dynamics (WBD) or highly dynamic trajectories on non-convex terrain. NLP, on the other hand, accommodates system nonlinearities and enables the exploration of richer, more realistic dynamic motions \cite{posa2014direct, dai2014whole, dafarra2022dynamic}. However, NLP is challenged by multiple local minima induced by terrain morphology and locomotion patterns, requiring careful initial guess selection to enhance the likelihood of finding optimal solutions. Solving the nonlinear equations in NLP can be computationally intensive and prone to numerical instability, potentially increasing solution times and limiting online applicability. Many NLP solvers rely on gradient-based methods requiring derivatives of the cost function and constraints. When applied to legged locomotion, these techniques demand differentiable contact descriptions. The method in \cite{winkler2018gait} implements an NLP problem that fixes the number of contact events while letting the solver determine step timings and footholds on a continuous heightmap. This approach avoids CCs but significantly limits the search space. Alternatively, \cite{jenelten2022tamols} uses onboard sensors to capture terrain and represent it through a non-smooth heightmap. Terrain gradients are computed via numerical differentiation, but solver accuracy is limited by map resolution. Filtered maps are used separately for foothold placement, torso orientation, and edge avoidance. The gravito-inertia acceleration (GIA) model is used (\Cref{table:comparison}).

Another methodology for nonlinear optimization is \emph{differential dynamic programming} (DDP). Traditionally it does not handle constraints explicitly, but extensions like Feasibility-driven DDP (FDDP \cite{mastalli2020crocoddyl} and BOX-FDDP \cite{mastalli2022feasibility}) manage infeasible initial guesses and control constraints. DDP can achieve greater computational efficiency compared to NLP, making DDP-based planners suitable for online applications (\Cref{table:comparison}).

In parallel to optimization-based methods, \emph{reinforcement learning} (RL) has emerged as a powerful tool for discovering robust control policies in environments with high variability and uncertainty but often requires extensive training data and may perform inefficiently without demonstration priors. \Cref{table:comparison} compares the state-of-the-art optimizers listed above.

\begin{figure}[!t]
    \centering
    \includegraphics[width=0.9\linewidth]{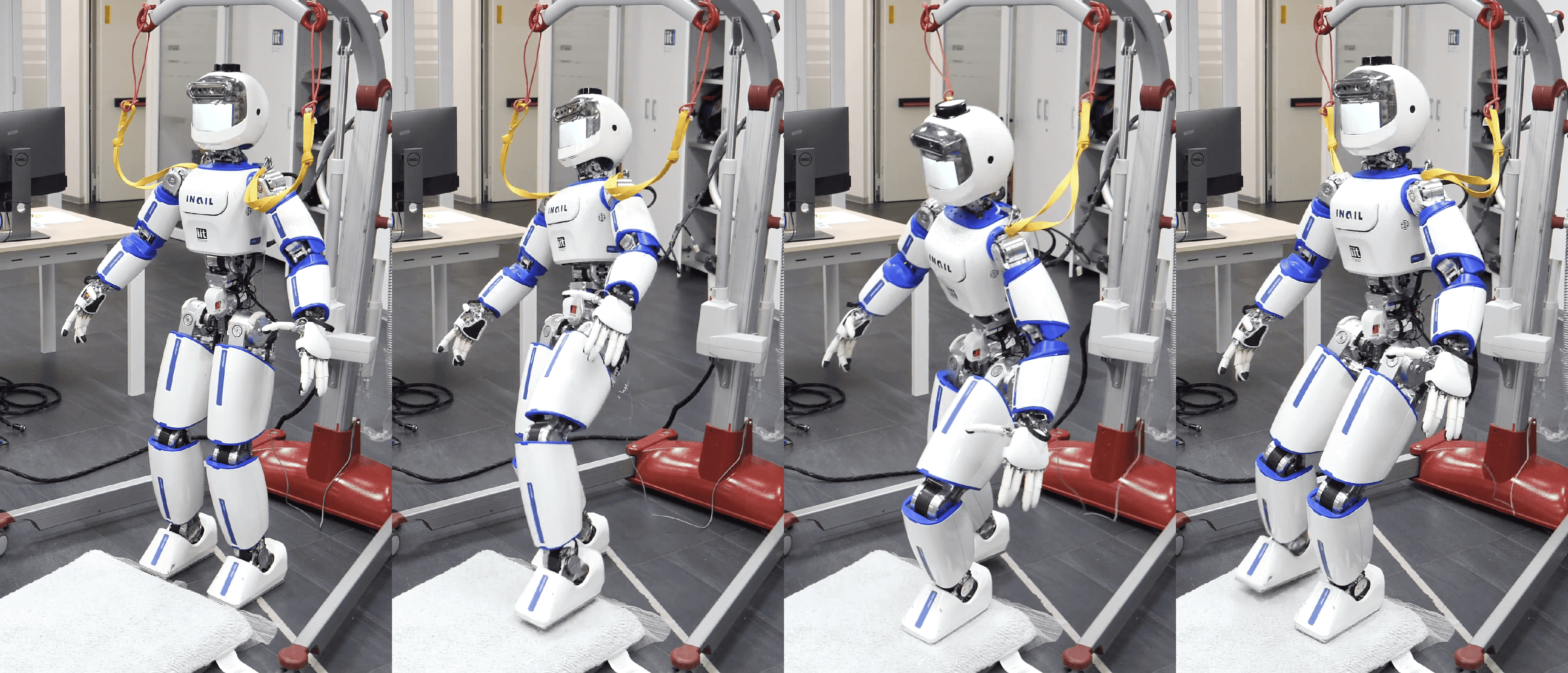}
    \caption{The robot ErgoCub performing a step-up trajectory (open~loop).}
    \label{fig:ergoCub}
\end{figure}

This letter tackles the challenge of humanoid motion planning on uneven terrain via a contact-implicit TO framework. Unlike prior approaches, our method integrates the robot WBD with a differentiable terrain and a mathematically coupled CC-based contact model. A notable aspect of our terrain model is its ability to remain smooth while closely approximating discontinuities common in human-centric environments. Our approach formulates planning as an optimal control (OC) problem solved via NLP. The key contributions are:
\begin{itemize}[align=right,itemindent=2em,leftmargin=0pt]
\item The introduction of a differentiable representation of uneven terrain, combined with generalized contact constraints.
\item A mass-scaling of the system's centroidal dynamics to improve efficiency and dynamic motion results.
\item The validation of our method by generating locomotion trajectories on non-flat surfaces, showcasing its ability to synthesize dynamic motions.
\end{itemize}
We conducted preliminary tests to evaluate the feasibility of the generated trajectories, validating the approach through both simulations and real-world experiments on the humanoid robot ErgoCub over non-coplanar terrain (\Cref{fig:ergoCub}).

\begin{table}[!t]
    \centering
    \caption{Comparison of optimizers for planning on uneven terrain.}
    \label{table:comparison}
    \renewcommand{\arraystretch}{1.2}
    \rowcolors{1}{}{gray!10}
    \begin{tabular}{rlllllllll}
    \toprule
    & \textbf{A} & \textbf{B} & \textbf{C} & \textbf{D} & \textbf{E} & \textbf{F} & \textbf{G} & \textbf{H} & \textbf{I}\\ 
    \midrule
    \textbf{Ace17 \cite{aceituno2017simultaneous}} & & \checkmark & & & CD & MICP & \checkmark & & \\
    \textbf{Win18 \cite{winkler2018gait}} & & \checkmark & \checkmark & \checkmark & CD & NLP & & \checkmark & \\
    \textbf{Mas20 \cite{mastalli2020crocoddyl}} & & \checkmark & & \checkmark & WBD & FDDP & \checkmark & \checkmark & \\ 
    \textbf{Ton20 \cite{tonneau2020sl1m}} & & \checkmark & & & CoM & MICP & \checkmark & \checkmark & \\ 
    \textbf{Jen22 \cite{jenelten2022tamols}} & & \checkmark & \checkmark & & GIA & NLP & \checkmark & & \checkmark \\ 
    \textbf{Our Work} & \checkmark & \checkmark & \checkmark & \checkmark & WBD & NLP & & \checkmark & \checkmark \\ 
    \bottomrule
    \end{tabular}
    \\[5pt]
    \footnotesize{A: \textit{differentiable terrain}; B: \textit{uneven surfaces}; C: \textit{strongly curved or inclined surfaces}; D: \textit{dynamic motions}; E: \textit{dynamic model}; F: \textit{optimization method}; G: \textit{online generation}; H: \textit{humanoid robot}; I: \textit{implicit contacts}.}
\end{table}

\section{Background}
\label{sec:background}
\subsection{Notation}
\label{sec:background:notation}
\begin{itemize}
    \item $n$-tuples in the real vector space are denoted by bold lowercase letters $\bm{x} \in \mathbb{R}^n$. $\bm{0}_n \in \mathbb{R}^n$ is the null $n$-tuple.

    \item Matrices are denoted by uppercase letters $M \in \mathbb{R}^{m \times n}$. $I_n, \, O_n \in \mathbb{R}^{n \times n}$ represent, respectively, the identity and the square null matrices of order $n$.

    \item Geometric points in the 3D Euclidean space are denoted by uppercase upright letters $\mathrm{P} \in \mathbb{E}^3$, while geometric vectors in the 3D Euclidean vector space are denoted by bold lowercase letters $\bm{v} \in \mathbb{\vec{E}}^3$.
    
    \item Reference frames are denoted by calligraphic letters $\mathcal{F}$. $\mathrm{O}_\mathcal{F}$ is the origin of $\mathcal{F}$, while $[\mathcal{F}]$ is its orientation.

    \item Given two frames $\mathcal{A}$ and $\mathcal{B}$, with $\mathcal{B} [\mathcal{A}]$ we refer to a new frame characterized by the same origin of $\mathcal{B}$ and the same orientation of $\mathcal{A}$ (mixed frames \cite{traversaro2019multibody}).
    
    \item Coordinates of points and vectors expressed with respect to (w.r.t.) a frame $\mathcal{F}$ are denoted by ${}^{\mathcal{F}}\mathrm{P},\, {}^{\mathcal{F}}\bm{v} \in \mathbb{R}^3$.
    
    \item $\mathcal{A}$ is the absolute inertial frame. Its origin is positioned at ground level and its $z$-axis points in the direction opposite to the local gravitational field.
    
    \item $\mathcal{B}$ is the body-fixed frame associated with the floating-base of the multi-body system (the waist of the robot).
    
    \item $\mathcal{C}_i$ is the frame associated with the $i$-th contact point $\mathrm{P}_i$ located on the foot contact point and oriented as $\mathcal{A}$.
    
    \item $\mathcal{G}$ is the centroidal frame \cite{saccon2017centroidal}, whose origin is the robot center of mass $\text{G}$, and oriented as $\mathcal{A}$.

    \item ${}^{\mathcal{A}}\mathrm{O}_{\mathcal{B}} \in \mathbb{R}^3$ is the translation between $\mathrm{O}_\mathcal{A}$ and $\mathrm{O}_\mathcal{B}$, ${}^{\mathcal{A}}R_{\mathcal{B}} \in \mathrm{SO}(3)$ is the rotation aligning $[\mathcal{A}]$ with $[\mathcal{B}]$, ${}^{\mathcal{A}}H_{\mathcal{B}} \in \mathrm{SE}(3)$ is the transformation between $\mathcal{A}$ and $\mathcal{B}$. 

    \item $S(\bm{x}): \mathbb{R}^3 \to \mathbb{R}^{3 \times 3}$ is the operator associating a 3-tuple $\bm{x}$ with a matrix $S$, such that $S(\bm{x}) \, \bm{y} \triangleq \bm{x}\times\bm{y}, \, \forall \bm{x},\, \bm{y} \in \mathbb{R}^3$, with $\times$ the usual cross-product in $\mathbb{R}^3$.
\end{itemize}

\subsection{Humanoid Robot Modeling}
\label{sec:background:humanoid_modeling}
A humanoid robot can be modeled as a multibody system consisting of $n + 1$ rigid bodies (\emph{links}) connected by means of $n$ \emph{joints} with one degree of freedom.
The robot \emph{configuration} is defined as the pair $q \triangleq ({}^{\mathcal{A}}H_{\mathcal{B}},\, \bm{s})$, which belongs to the Lie group $\mathrm{SE}(3) \times \mathbb{R}^n$, where $\bm{s} \in \mathbb{R}^n$ is the joint position vector. As $q$ is not a vector, we refrain from using boldface notation. According to Lie group theory, we can define the vector of the system \emph{generalized velocities} $\bm{\nu}\Tr \! 
\triangleq [{}^{\mathcal{B} [\mathcal{A}]} \mathbf{v}\Tr_{\mathcal{A},\, \mathcal{B}} \quad \bm{v}_s\Tr] \in \mathbb{R}^{6+n}$, where $\mathbf{v}_{\mathcal{A},\, \mathcal{B}}$ is the twist describing the instantaneous relative motion of $\mathcal{B}$ w.r.t. $\mathcal{A}$, while $\bm{v}_s$ are the joint velocities. 

The floating-base system dynamics is completely described by the Euler-Poincar\'e equations, here expressed in the centroidal frame \cite[Eq. (3.95b)]{traversaro2017thesis}, \cite{garofalo2015inertially}:
\begin{equation}
    \label{eq:euler_poincare}
    {}^{\mathcal{G}}M(q) \, \dot{\bm{\nu}} + {}^{\mathcal{G}}\bm{b}(q, \bm{\nu}) = 
    \begin{bmatrix}
        \bm{0}_6 \\
        \bm{\tau}_s
    \end{bmatrix} +
    \sum_{i = 1}^{n_c} \, {}^{\mathcal{C}_i}\!J_i\Tr\!(q) \, {}^{\mathcal{C}_i}\mathbf{f}_i,
\end{equation}
where ${}^{\mathcal{G}}M \in \mathbb{R}^{(6+n) \times (6+n)}$ is the generalized centroidal inertia matrix, ${}^{\mathcal{G}}\bm{b} \in \mathbb{R}^{6+n}$ is the bias term which accounts for the Coriolis, centrifugal, and gravitational effects, $\bm{\tau}_s \in \mathbb{R}^n$ are the internal joint torques, and ${}^{\mathcal{C}_i}\mathbf{f}_i \in \mathbb{R}^6$ is the $i$-th external wrench expressed w.r.t. $\mathcal{C}_i$. The transpose of the Jacobian matrix ${}^{\mathcal{C}_i}J_i\Tr\!(q) \in \mathbb{R}^{(6 + n) \times 6}$ maps the $i$-th external wrench to a vector of external \emph{generalized forces}.

\subsection{Centroidal Momentum Dynamics}
\label{sec:background:centroidal_dynamics}
Trajectory planning for floating-base multi-body systems can be computationally demanding due to the high dimensionality and nonlinearities of the full dynamic model. A widely adopted simplification relies on CD \cite{orin2013centroidal}, which encapsulates the fundamental characteristics of whole-body motion, such as momentum and contact interactions, while substantially reducing the dimensionality of the state space:
\begin{equation}
    {}^{\mathcal{G}}\dot{\mathbf{h}} 
    = 
    \begin{bmatrix}
        m \, {}^{\mathcal{G}}\bm{g} \\
        \bm{0}_3
    \end{bmatrix} 
    +
    \sum_{i = 1}^{n_c}
    \begin{bmatrix}
        I_3 & O_3 \\
        S({}^{\mathcal{G}}\mathrm{O}_{\mathcal{C}_i}) & I_3
    \end{bmatrix}
    {}^{\mathcal{C}_i}\mathbf{f}_i,
\end{equation}
where ${}^{\mathcal{G}}\mathbf{h} \in \mathbb{R}^6$ is the \emph{centroidal momentum} \cite{orin2013centroidal}, $\bm{g} \in \vec{\mathbb{E}}^3$ is the local gravity acceleration, $m \in \mathbb{R}$ is the mass of the system, and $\mathrm{O}_{\mathcal{C}_i} \in \mathbb{E}^3$ is the $i$-th contact point. It describes how the system's linear and angular momenta evolve, ignoring its internal structure. It is completely equivalent to the first six rows of \Cref{eq:euler_poincare} (i.e., the projection of the full dynamics onto the 6D space of the centroidal momentum). 
The \emph{centroidal momentum matrix} \cite{orin2013centroidal} represents the linear map between $\bm{\nu}$ and ${}^{\mathcal{G}}\mathbf{h}$, denoted by the matrix ${}^{\mathcal{G}}J_\text{CMM} \in \mathbb{R}^{6 \times (6+n)}$:
\begin{equation}
\label{eq:centroidal_momentum_matrix}
    {}^{\mathcal{G}}\mathbf{h} = {}^{\mathcal{G}}J_\text{CMM}(q) \, \bm{\nu}.
\end{equation}

\subsection{Optimal Control Problem and Nonlinear Programming}
\label{sec:background:optimal_control}
A constrained OC problem involves finding a control policy for a \emph{dynamical system} to minimize a cost functional $\mathfrak{J}$ and satisfying constraints involving state and control time signals. NLP techniques are crucial for addressing the optimization challenges of OC problems, since the former represent a finite-dimensional approximation of the latter. An NLP problem consists of three main elements:
\begin{itemize}
    \item a $p$-tuple of \emph{optimization variables} $\bm{\chi} \in \mathbb{R}^p$, that can be freely chosen within its domain.
    \item A scalar function in the optimization variables $\Gamma(\bm{\chi}): \mathbb{R}^p \to \mathbb{R}$ to be minimized (\emph{cost function}).
    \item A set of $k$ equations or inequalities that algebraically link the optimization variables $g_i(\bm{\chi}) \geq 0$ (\emph{constraints}).
\end{itemize}
The solution of the NLP problem is the vector of optimization variables $\bm{\chi}^\star$ that simultaneously satisfies all constraints and minimizes the objective function:
\begin{equation*}
\begin{gathered}
    \bm{\chi}^\star \triangleq \underset{\bm{\chi}\, \in \, \Omega}{\arg \min} \; \Gamma(\bm{\chi}).
\end{gathered}
\end{equation*}
The set $\Omega \triangleq \{ \bm{\chi} \in \mathbb{R}^p \! : \: g_i(\bm{\chi}) \geq 0,\, \forall i\}$ is called \emph{feasible set}. $\Gamma (\bm{\chi})$ and $g_i(\bm{\chi})$ are assumed to be differentiable functions. The differentiability of all the problem functions enables the use of derivative-based optimization algorithms. 
This includes gradient-based methods, which directly compute gradients and Hessian matrices, and interior point methods, which indirectly use derivatives through the formulation and manipulation of barrier functions \cite{bomze2010interior}. 

The infinite-dimensional OC problem can be transcribed into a finite-dimensional NLP problem by using the so-called \emph{direct methods} (e.g., single shooting, multiple shooting or direct collocation \cite{diehl2006fast}), and can be easily tackled through the use of numerical strategies \cite{nocedal1999numerical, betts2010practical}.

\section{Contributions}
\label{sec:contributions}
\subsection{Uneven Terrain Modeling}
\label{sec:contributions:terrain_modeling}
\begin{figure}[!t]
    \vspace{2mm}
    \centering
    \includegraphics[width=0.85\linewidth]{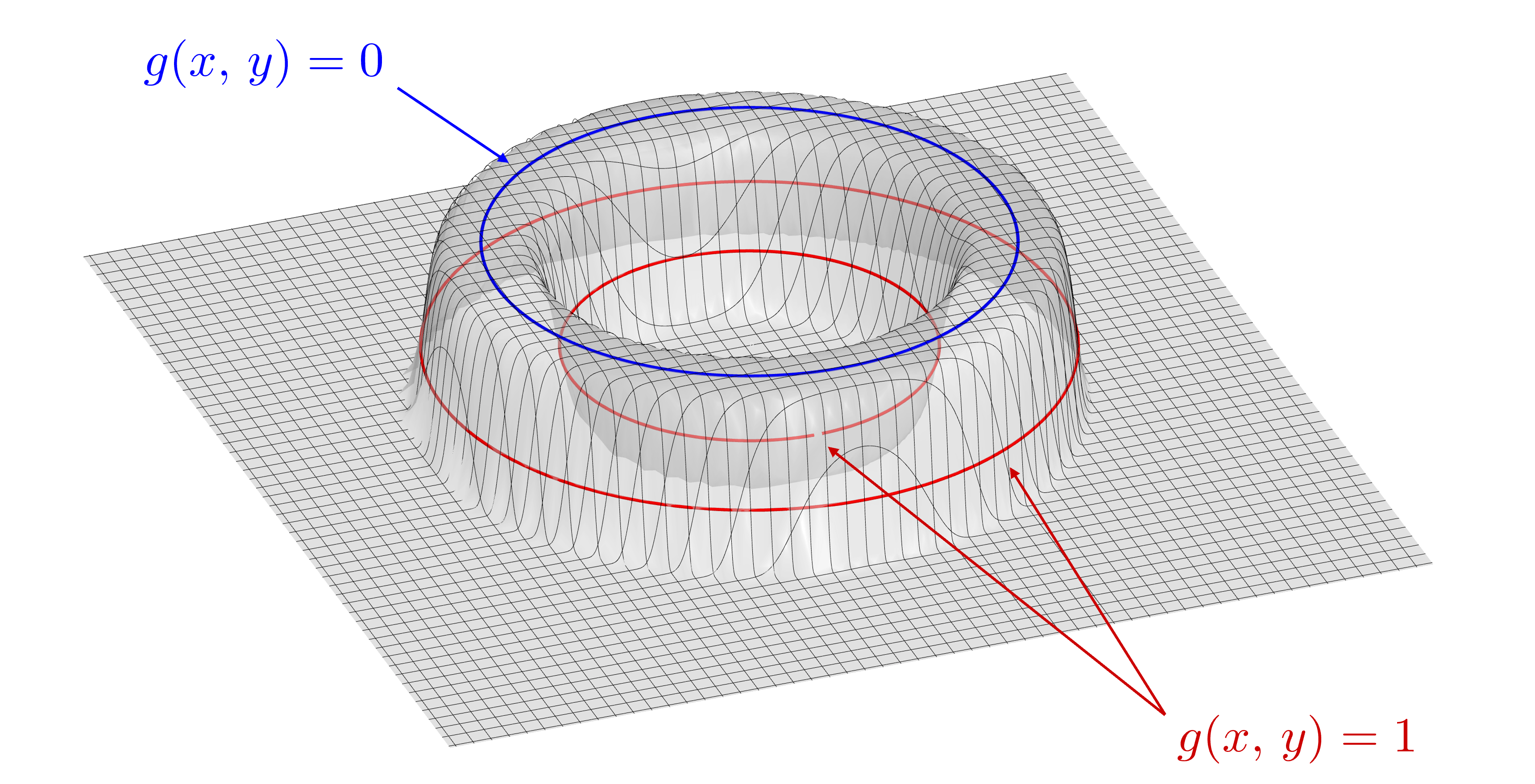}
    \vspace{-2mm}
    \caption{Surface with a torus-like shape: $f(x,\, y) = \exp \left \{ -g(x,\, y)^s \right \}$ with $g(x,\, y) = (x^2 + y^2 - 2)^2$ and $s = 4$. The blue curve is the geometric locus of points satisfying the implicit function $g(\bm{x}) = 0$, the red curves are the geometric locus of points satisfying the implicit function $g(\bm{x}) = 1$.}
    \label{fig:torus2}
\end{figure}

This section introduces a terrain model that \emph{smoothly} approximates discontinuous shapes like steps and ramps. The model ensures a \emph{differentiable} representation, crucial for optimization methods relying on derivatives. To achieve smoothness while capturing discontinuities, each terrain \emph{element} is modeled as a 2D differentiable manifold, defined by the graph of the following smooth map $f(\bm{x}):\mathbb{R}^2 \to \mathbb{R}$:
\begin{equation}
    \label{terrain_function}
    f(\bm{x}) \triangleq \exp \left (-g(\bm{x})^{s} \right ) \, \pi(\bm{x}),
\end{equation}
where $\pi(\bm{x}):\mathbb{R}^2 \to \mathbb{R}$ is the explicit function describing a generally oriented plane in the 3D space, $g(\bm{x}):\mathbb{R}^2 \to \mathbb{R}$ is a positive semi-definite differentiable function and $s \geq 2$ is a real number, called \emph{sharpness}. To clarify the steps that led to the definition of such a function, we can start with the definition of the one-dimensional parametric function:
\begin{equation*}
f(\xi) \triangleq \exp \left ( -\xi^{\, s} \right ),
\end{equation*}
with $\xi \in [0,\, +\infty) \subset \mathbb{R}$ and $s \in [2,\, +\infty) \subset \mathbb{R}$. For each value of the sharpness $s$ in its domain, there are two fixed points in the function graph (they do not depend on $s$):
\begin{equation*}
    f(0) = 1 \quad \text{and} \quad f(1) = e^{-1} .
\end{equation*}
When $s \to +\infty$, it approximates a step function. The idea is to make the variable $\xi$ a function of the planar variables $\bm{x} \in \mathbb{R}^2$ ($\xi = g(\bm{x})$), in order to obtain surfaces with a ``prism"-like shape. This means that $f(\bm{x})$ is nearly zero everywhere except in a limited region where it is close to one. (\Cref{fig:torus2}):
\begin{equation*}
    f(\xi) \triangleq \exp \left ( -\xi^{\, s} \right ) \: \to \: f(g(\bm{x})) \triangleq \exp \left ( -g(\bm{x})^s \right ).
\end{equation*}
The function $g(\bm{x})$ should be chosen in such a way that:
\begin{itemize}
    \item $g(\bm{x}) \geq 0 \quad \forall \bm{x} \in \mathbb{R}^2$,
    \item $g(\bm{x}) = 0$ is the implicit function that describes the shape of the upper part of the ``prism" (usually a point),
    \item $g(\bm{x}) = 1$ is the implicit function that describes the shape of the lower part of the ``prism".
\end{itemize}
This geometric model allows us to represent elements such as parallelepipeds, cylinders, tori, and ``prisms" with a desired base shape. Here are some examples of common shapes:
\begin{itemize}[align=right,itemindent=2em,leftmargin=0pt]
\item \emph{Rectangle}:
\begin{equation*}
g(x,\, y) \triangleq \left | x / w_x \right |^{p} + \; \left | y / w_y \right |^{p} \!,
\end{equation*}
where $p \in \mathbb{R}^{+}$ is a parameter related to the sharpness of the shape corners and $w_x,\, w_y \in \mathbb{R}^{+}$ are the half-widths of the rectangle edges. If $w_x = w_y$ the shape is a \emph{square}. If $p = 2$ the shape is an \emph{ellipse} or a \emph{circle} (if $w_x = w_y$).
\item \emph{Equilateral Triangle}:
\begin{equation*}
g(x,\, y) \triangleq \sum _{k=1}^3 \left | x \cos \left (\tfrac{2}{3} \pi \, k \right ) + y \sin \left (\tfrac{2}{3} \pi \, k \right ) - \tfrac{1}{2} \right |^p \!.
\end{equation*}
\end{itemize}
If we want a ``prism" with an inclined upper base, we can multiply the exponential function by the expression of a generically oriented plane:
\begin{equation*}
\pi(\bm{x}) \triangleq \bm{a}\Tr \bm{x} + h,
\end{equation*}
where $\bm{a} \in \mathbb{R}^2$ is the $2$-tuple collecting the angular coefficients of the plane and $h \in \mathbb{R}$ is the height of the plane in $\bm{x} = \bm{0}_2$. Where the exponential function is close to 1, the surface will approximate that plane. Thus, we have obtained \Cref{terrain_function} introduced earlier. Eventually, to obtain even more complex shapes, it is possible to replace the plane $\pi(\bm{x})$ with a generic function $h(\bm{x})$ (\Cref{fig:trajectories}). In the end, we can make an affine change of coordinates to translate (with $\bm{b}$), rotate and/or deform (with $A$) the original shape: $\bm{x} \triangleq A \, \tilde{\bm{x}} + \bm{b}$.

Individual elements $f_i(\bm{x})$ can be combined through a simple addition to create more complex shapes (e.g. stairs or inclined jumping platforms) $F(\bm{x}) \triangleq \sum^{n_f}_{i = 1} f_i(\bm{x})$.
Given the explicit function of a smooth surface $z = F(\bm{x})$, it is always possible to evaluate the tangent plane at any point $Q$ and the vector perpendicular to this (\emph{local surface normal}):
\begin{equation}
\label{eq:normals}
{}^{\mathcal{A}}\bm{n}_\mathrm{Q} \triangleq 
\begin{bmatrix}
    \left. -\nabla F(\bm{x}) \right|_{\bm{x}={}^{\mathcal{A}}\mathrm{Q}} & 1
\end{bmatrix}\Tr \! \in \mathbb{R}^3.
\end{equation}
This vector will play a pivotal role in the contact modeling.

\subsection{Contact Modeling on Smooth Uneven Terrain}
\label{sec:contribution:contact_model}
We consider a point contact model with Coulomb friction, requiring a tangent plane and normal vector at each surface point, as ensured by differentiable surfaces. Assuming a known terrain, we can derive a vector field that assigns the local normal vector $\bm{n}_{\mathrm{Q}}$ to each point $\mathrm{Q}$. Let us define:
\begin{itemize}
\item the orthogonal projector on the plane perpendicular to a given vector $\bm{x}$. It is defined as $P_\perp(\bm{x}) \triangleq -S(\hat{\bm{x}})^2$,

\item the orthogonal projector along the axis parallel to a given vector $\bm{x}$. It is defined as $P_\paral(\bm{x}) \triangleq I_3 + S(\hat{\bm{x}})^2$,
\end{itemize}
where $\hat{\bm{x}} \triangleq \bm{x} / \| \bm{x} \| \in \mathbb{R}^3$, with $\| \cdot \|$ the Euclidean norm.

Before providing the constraints that formally describe contacts on uneven terrain, it is important to emphasize the assumptions behind this contact model: contacts are assumed to be rigid (neither body deforms upon contact) and static (no slipping), whereas impact events are assumed to be inelastic (kinetic energy is generally not conserved). 

With $\bm{f}_\mathrm{P}$ we denote a contact force applied to the point $\mathrm{P}$, while with $\bm{d}_\mathrm{P} \triangleq (\mathrm{P} - \mathrm{Q}) \in \vec{\mathbb{E}}^3$ we denote the distance vector connecting the point on the surface $\mathrm{Q}$ to the point $\mathrm{P}$, fixed to the foot. $\mathrm{Q}$ is chosen as the point located at the intersection of the surface with the vertical line passing through the point $\mathrm{P}$. Alternative selection criteria may also be considered. It is imperative, however, that $\mathrm{Q} \equiv \mathrm{P}$ during the contact phase. The coefficient of static friction is represented by $\mu$.
\begin{itemize}
\item \emph{Contact Condition:} 
\begin{equation}
\label{eq:complem_condition}
{}^{\mathcal{A}}\bm{d}_\mathrm{P}\Tr P_\paral({}^{\mathcal{A}}\bm{n}_\mathrm{Q}) \, {}^{\mathcal{A}}\bm{f}_\mathrm{P} = 0,
\end{equation}
\item \emph{No-slip Condition:} 
\begin{equation}
\label{eq:no_slip_condition}
P_\perp({}^{\mathcal{A}}\bm{n}_\mathrm{Q}) \, {}^{\mathcal{A}}\bm{v}_\mathrm{P} \, {}^{\mathcal{A}}\bm{f}_\mathrm{P}\Tr \! = O_3,
\end{equation}
\item \emph{Friction Cone Condition:}
\begin{equation}
\label{eq:friction_cone_condition}
{}^{\mathcal{A}}\bm{f}_\mathrm{P}\Tr \! \left [ \mu^2 \, P_\paral({}^{\mathcal{A}}\bm{n}_\mathrm{Q}) - P_\perp({}^{\mathcal{A}}\bm{n}_\mathrm{Q}) \right ] {}^{\mathcal{A}}\bm{f}_\mathrm{P} \geq 0,
\end{equation}
\item \emph{No-penetration Condition:} 
\begin{equation}
\label{eq:no_penetration_condition}
{}^{\mathcal{A}}\bm{n}_\mathrm{Q}\Tr \, {}^{\mathcal{A}}\bm{d}_\mathrm{P} \geq 0,
\end{equation}
\item \emph{Unilateral Contact Condition:} 
\begin{equation}
\label{eq:unilateral_contact_condition}
{}^{\mathcal{A}}\bm{n}_\mathrm{Q}\Tr \, {}^{\mathcal{A}}\bm{f}_\mathrm{P} \geq 0.
\end{equation}
\end{itemize}
The CCs expressed by the \Cref{eq:complem_condition,eq:no_slip_condition} are implemented in their \emph{relaxed} versions (i.e., $\varepsilon_{\min} \leq \text{CC} \leq \varepsilon_{\max}$, instead of $\text{CC} = 0$. $[\varepsilon_{\min},\, \varepsilon_{\max}]$ is the desired relaxation interval), since this mitigates issues of numerical instability, as detailed in \cite{manchester2019contact}. Thus, all the conditions are described by inequalities.

\subsection{OC Problem - Dynamical System}
\label{sec:contribution:dynamical_system} 
In our case, state $\bm{x}$ and control $\bm{u}$ are defined as:
\begin{equation*}
\begin{aligned}
\bm{x} &\triangleq 
\begin{bmatrix}
{}^{\mathcal{G}}\mathbf{h}\Tr &
{}^{\mathcal{A}}\mathrm{O}\Tr_{\mathcal{G}} &
{}^{\mathcal{A}}q\Tr_{\mathcal{B}} &
{}^{\mathcal{A}}\mathrm{P}\Tr_i &
{}^{\mathcal{A}}\bm{f}\Tr_i
\end{bmatrix}\Tr \!\!,
\\
\bm{u} &\triangleq 
\begin{bmatrix}
{}^{\phantom{\mathcal{G}}}\bm{\nu}\Tr &
{}^{\mathcal{A}}\bm{u}\Tr_{\mathrm{P},\, i} &
{}^{\mathcal{A}}\bm{u}\Tr_{\bm{f},\, i}
\end{bmatrix}\Tr \!\!,
\end{aligned}
\end{equation*}
where $\mathrm{P}_i \in \mathbb{E}^3$ and $\bm{f}_i \in \vec{\mathbb{E}}^3$ represent, respectively, the contact points and the associated contact forces on the robot feet. The quantities $\bm{u}_{\mathrm{P},\, i} \text{ and } \bm{u}_{f,\, i} \in \vec{\mathbb{E}}^3$ are system inputs which we assume the planner can control in order to generate smoother variations in contact point positions and forces. These inputs allow for finer control over the contact dynamics, ensuring more gradual and smooth transitions in the system behavior.
The robot configuration ${}^{\mathcal{A}}q_{\mathcal{B}}$ and velocity $\bm{\nu}$ are defined as:
\begin{equation*}
\begin{aligned}
{}^{\mathcal{A}}q_{\mathcal{B}}\Tr \! 
&\triangleq 
\begin{bmatrix}
{}^{\mathcal{A}}\mathrm{O}\Tr_{\mathcal{B}} &
{}^{\mathcal{A}}\rho\Tr_{\mathcal{B}} &
\bm{s}\Tr 
\end{bmatrix} \in \mathbb{R}^{3} \times \mathrm{U}(\mathbb{H}) \times \mathbb{R}^{n}, 
\\[0.5em]
\bm{\nu}\Tr \!
&\triangleq
\begin{bmatrix}
{}^{\mathcal{B}[\mathcal{A}]} \bm{v}\Tr_{\mathcal{A},\, \mathcal{B}} & {}^{\mathcal{B}[\mathcal{A}]} \bm{\omega}\Tr_{\mathcal{A},\, \mathcal{B}} & \bm{v}_s\Tr
\end{bmatrix}
\in \mathbb{R}^{6 + n},
\end{aligned}
\end{equation*}
where ${}^{\mathcal{A}}\rho_{\mathcal{B}} \in \mathrm{U}(\mathbb{H})$ represents the floating-base orientation w.r.t. the absolute frame in terms of unit quaternions\footnote{Rotation matrices are parameterized using unit quaternions ${}^{\mathcal{A}}\rho_{\mathcal{B}} \in \mathrm{U}(\mathbb{H})$ by means of the formula ${}^{\mathcal{A}}R_{\mathcal{B}} = \left [ \rho_s \, I_3 + S(\rho_v) \right ]^2 + \rho_v \, \rho_v\Tr$, where $\rho_s \in \mathbb{R}$ is the scalar part, while $\rho_v \in \mathbb{R}^3$ is the vector part of ${}^{\mathcal{A}}\rho_{\mathcal{B}}$.}, while ${}^{\mathcal{B}[\mathcal{A}]} \bm{v}_{\mathcal{A},\, \mathcal{B}}$ and ${}^{\mathcal{B}[\mathcal{A}]} \bm{\omega}_{\mathcal{A},\, \mathcal{B}} \in \mathbb{R}^3$ represent, respectively, the relative linear and angular velocity of the floating-base w.r.t. the absolute frame, expressed w.r.t. the mixed frame $\mathcal{B}[\mathcal{A}]$.

We consider the combination of the centroidal dynamics:
\begin{equation*}
\label{eq:centroidal_dynamics}
{}^{\mathcal{G}}\dot{\mathbf{h}} =
\sum_{i = 1}^{n_c} 
\begin{bmatrix}
    I_3 \\
    S({}^{\mathcal{G}}\mathrm{P}_i)
\end{bmatrix}
{}^{\mathcal{C}_i}\bm{f}_i 
+
\begin{bmatrix}
    m \, {}^{\mathcal{G}}\bm{g} \\
    \bm{0}_3
\end{bmatrix} \quad \text{and} \quad
{}^{\mathcal{A}}\dot{\mathrm{O}}_{\mathcal{G}} = {}^{\mathcal{G}}\mathbf{h}_p / m,
\end{equation*}
where $\mathbf{h}_p$ is the linear momentum, and the full kinematics:
\begin{equation*}
\label{eq:full_kinematics}
{}^{\mathcal{A}}\dot{\mathrm{O}}_{\mathcal{B}} = {}^{\mathcal{B} [\mathcal{A}]}\bm{v}_{\mathcal{A},\, \mathcal{B}}, \quad
{}^{\mathcal{A}}\dot{\rho}_{\mathcal{B}} = {}^{\mathcal{B}[\mathcal{A}]} \bm{\omega}_{\mathcal{A},\, \mathcal{B}} \otimes \frac{1}{2} \, {}^{\mathcal{A}}\rho_{\mathcal{B}}, \quad \dot{\bm{s}} = \bm{v}_s,
\end{equation*}
where the operator $\otimes$ represents the Hamilton product. With an abuse of notation, ${}^{\mathcal{B}[\mathcal{A}]} \bm{\omega}_{\mathcal{A},\, \mathcal{B}}$ here denotes a $4$-tuple representing the components of a pure quaternion.
First-order dynamics have been applied to contact points and forces:
\begin{equation*}
\label{eq:contact_dynamics}
{}^{\mathcal{A}}\dot{\mathrm{P}}_i = {}^{\mathcal{A}}\bm{u}_{\mathrm{P},\, i} 
\quad \text{and} \quad
{}^{\mathcal{A}}\dot{\bm{f}}_i = {}^{\mathcal{A}}\bm{u}_{f,\, i}.
\end{equation*}

\emph{Scaling of the System Dynamics:} We observed that the planner converges more quickly to a feasible solution when the centroidal dynamics is independent of the robot total mass $m$. This is achieved by scaling the quantities ${}^{\mathcal{G}}\mathbf{h}$, ${}^{\mathcal{A}}\bm{f}_i$, and ${}^{\mathcal{A}}\bm{u}_{f,\, i}$ by $m$, resulting in mass-normalized quantities denoted with a tilde: ${}^{\mathcal{G}}\tilde{\mathbf{h}}$, ${}^{\mathcal{A}}\tilde{\bm{f}}_i$, and ${}^{\mathcal{A}}\tilde{\bm{u}}_{f,\, i}$. With these variables, the only equation where mass explicitly appears is the centroidal momentum consistency constraint (\Cref{sec:contributions:constraints}).

\subsection{OC Problem - Constraints}
\label{sec:contributions:constraints}
\subsubsection{Differential Constraints}
They consist of the mass-scaled equations presented in \Cref{sec:contribution:dynamical_system}. The related initial and final conditions are included in the equality constraints.
\subsubsection{Inequality Constraints}
\begin{itemize}[align=right,itemindent=2em,leftmargin=0pt]
\item \emph{Contact Constraints:} 
Inequalities introduced in \Cref{eq:complem_condition,eq:no_slip_condition,eq:friction_cone_condition,eq:no_penetration_condition,eq:unilateral_contact_condition}. On each foot, four contact points $\mathrm{P}_i$ have been chosen and correspond to the vertices of the sole ($n_c = 8$). Once set, their positions relative to the foot frame do not change during the motion (see point 3-\emph{Equality Constraints}).
\item \emph{Joint Limits:} $\bm{s}_\text{min} \leq \, \bm{s} \leq \bm{s}_\text{max}$, $\bm{v}_{s,\, \text{min}} \leq \, \bm{v}_s \leq \bm{v}_{s,\, \text{max}}$, and $\bm{\tau}_{s,\, \text{min}} \leq \, \bm{\tau}_s \leq \bm{\tau}_{s,\, \text{max}}$, with $\bm{\tau}_s$ evaluated inverting the Euler-Poincar\'e \Cref{eq:euler_poincare}. Joint torques can also be expressed on the motor side (that is, upstream of gearboxes), as presented in \cite{nava2018exploiting}, and are denoted as motor torques $\bm{\tau}_m$. In this way, it is possible to impose constraints directly on the maximum torques exerted by the motors, available on datasheets.
\end{itemize}
\subsubsection{Equality Constraints}
\begin{itemize}[align=right,itemindent=2em,leftmargin=0pt]
\item \emph{Rigid Connection of the Contact Points to the Feet, Unit Quaternions \emph{and} Center of Mass Consistency:} The constraints adopted here are consistent with those formally introduced by Dafarra \emph{et al.} \cite{dafarra2022dynamic}.
\item \emph{Centroidal Momentum Consistency:} It consists in the \emph{mass-scaled} version of \Cref{eq:centroidal_momentum_matrix}. This is the only equation in which the robot total mass appears explicitly.
\end{itemize}
\subsection{OC Problem - Tasks}
\label{sec:contributions:tasks}
We define a \emph{task} as a function whose minima represent an objective we aim to achieve and which represents a characteristic we want to be present in the generated motion. The weighted sum of multiple tasks constitutes the cost function. This allows mathematical encoding of high-level directives on the behavior that the system should exhibit. Bar-superscripted symbols denote constant desired values, while the subscript $W$ denotes the weighting matrices in the weighted norms.
\begin{itemize}[align=right,itemindent=2em,leftmargin=0pt]
\item \emph{State \emph{and} Control Regularization Tasks:}
\begin{equation}
\label{eq:regularization_task}
\Gamma \triangleq \left \| \bm{x} - \bar{\bm{x}} \right \|^2_{W_x}
\quad \text{and} \quad
\Gamma \triangleq \left \| \bm{u} - \bar{\bm{u}} \right \|^2_{W_u}.
\end{equation}
In this case, the regularization value $\bar{\bm{u}}$ is the null tuple. To evaluate the \emph{distance} between two unit quaternions $\rho,\, \bar{\rho}$, the \emph{pseudometric} $\phi(\rho,\, \bar{\rho}) \triangleq \min \{ \|\rho - \bar{\rho} \|,\, \| \rho + \bar{\rho} \| \}$ is used \cite{huynh2009metrics}.
\item \emph{Swing Height Task \emph{and} Force Regularization Task:} 
The tasks utilized herein correspond to those delineated in the framework proposed by Dafarra \emph{et al.} \cite{dafarra2022dynamic}.
\item \emph{Motor Torque Minimization Task:}
\begin{equation}
\label{eq:torque_task}
\Gamma \triangleq \left \| \bm{\tau}_m \right \|^2_{W_\tau}.
\end{equation}
It prevents the use of maximum available motor torques.
\item \emph{Lateral Foot Distance Task:}
\begin{equation}
\label{eq:lateral_distance_task}
\Gamma \triangleq \left | \left ( {}^{\mathcal{A}}\mathrm{O}_{\mathcal{F}_{\text{l}}} - {}^{\mathcal{A}}\mathrm{O}_{\mathcal{F}_{\text{r}}} \right )\!\!\Tr \; {}^{\mathcal{A}} \bm{j}_{\mathcal{A}} - \bar{d}_y \right |^2 ,
\end{equation}
where $\bar{d}_y \in \mathbb{R}$ is a constant value that specifies the desired lateral foot distance. It prevents leg self-penetration during walking motions along the $\bm{i}_\mathcal{A}$ direction.
\end{itemize}

\subsection{Choice of the Initial Guess}
\label{sec:contribution:initial_guess}
Given the non-convex nature of the NLP problem, the choice of the initial guess plays a crucial role in shaping the resulting trajectories, as different choices may lead to distinct \emph{local minima}. This work highlights the importance of the initial guess for the time evolution of contact forces. For instance, where these forces are set equal to zero, the computed solution results in realistic jumping motions. The initial guess is built by interpolating between two robot configurations (initial and final) constrained to static equilibrium.
\subsubsection{Robot Configurations in Static Equilibrium}
\label{sec:contribution:static_configurations}
This subsection addresses the challenge of determining the desired configuration of a robot and the contact forces exerted upon it when positioned on uneven terrain. The objective is to compute the robot configuration, ${}^{\mathcal{A}}q_{\mathcal{B}}$, given the CoM position, ${}^{\mathcal{A}}\mathrm{O}_\mathcal{G}$, and feet poses, ${}^{\mathcal{A}}H_{\mathcal{F}_{\text{l}}}$ and ${}^{\mathcal{A}}H_{\mathcal{F}_{\text{r}}}$, ensuring that each foot maintains contact at no fewer than three distinct points.

Our approach follows a systematic procedure for each foot:
\begin{itemize}[align=right,itemindent=2em,leftmargin=0pt]
    \item Select the desired 2D pose, defined by the planar coordinates of a contact point and the yaw angle $\psi$.
    \item Determine the height of this contact point to ensure its alignment with the ground surface.
    \item Rotate the foot about this initial contact point, $\text{P}_0$, by adjusting roll $\varphi$ and pitch $\vartheta$ angles so that two additional contact points, $\text{P}_1$ and $\text{P}_2$, also lie on the ground surface.
\end{itemize}
Since the couple $(\varphi, \vartheta)$ is generally non-unique, the solution is determined by minimizing $\Gamma \triangleq \left \| \varphi \right \|^2 + \left \| \vartheta \right \|^2$ to ensure least deviations in the ankle angles. The problem is constrained by the condition that $\text{P}_1$ and $\text{P}_2$ belong to the ground surface.
The contact points' positions are then used to solve a statics problem to evaluate feasible contact forces. This problem is hyper-static, meaning static equilibrium equations alone can't determine the contact forces uniquely. Therefore, a criterion must be defined to select one feasible solution from the infinite possibilities. This criterion can be framed as an optimization problem with contact forces as variables, subject to contact constraints \Cref{eq:complem_condition,eq:no_slip_condition,eq:friction_cone_condition,eq:unilateral_contact_condition,eq:no_penetration_condition}.

Once feasible contact forces are evaluated, we seek a robot configuration ${}^{\mathcal{A}}q_{\mathcal{B}}$ with the CoM in the desired position and feet in the desired poses. This problem can also be framed as an optimization problem, where the variables are the components of ${}^{\mathcal{A}}q_{\mathcal{B}}$. Constraints include the desired foot poses, the CoM position, joint limits, and unit norm for the base quaternion. The objective function is a regularization task on the joint positions and base pose.
\subsubsection{Interpolation between Two States}
A continuous transition between initial and final states is done using \emph{linear interpolation} for the vector component and \emph{spherical linear interpolation} for the quaternion component. During jumping, contact forces are set to zero for the flight phase. For walking, contact forces on the feet are alternately nullified. These guesses are simple and rough but sufficiently expressive to guide algorithmic convergence towards the desired behaviors.

\section{Results}
\label{sec:results}
This section presents the results obtained by solving the discussed OC problem, taking into account all the details introduced in the previous sections. The transcription uses the multiple shooting method \cite{betts2010practical}. State-transition equality constraints are derived by discretizing the differential equations (\Cref{sec:contribution:dynamical_system}) using the \emph{explicit Euler} method. To preserve the structure of the group of unit quaternions, an integration over the Lie algebra followed by the application of the \emph{exponential map} is used \cite{sola2018micro}. The integration step $\Delta t$ is assumed to be constant. The \emph{time horizon} $[t_i,\, t_f] \subset \mathbb{R}$ is finite and its length $T$ has been varied according to the required task. It has been divided into $N$ \emph{time intervals} of the same length $\Delta t = T/N$. The NLP problem is solved using the software package \texttt{IPOPT} \cite{wachter2006implementation}, employing the \texttt{mumps} and \texttt{MA57} linear solvers \cite{hsl2007collection}. The pipeline from the problem definition to its solution is implemented entirely in \texttt{Python~3.10} with the help of the symbolic framework \texttt{CasADi} \cite{andersson2019casadi}.
The dimension of the search space we considered is $88 \, (N + 1) + 78 \, N$ and the optimization computation time ranges from tens of seconds to a few minutes on a common laptop (11\textsuperscript{th} Gen Intel Core i7, 32 GB of RAM). The selected weights of the cost function terms are wisely crafted based on the desired behavior. For some motions that include jumps, the weight of the task \eqref{eq:torque_task} has been significantly decreased and the torque limits have been deactivated. The proposed framework was tested using the dynamical models of two different robots. The transition between models required minimal changes to the planner, highlighting its robustness to parameter variations.

\subsection{Generated Trajectories}
\label{sec:results:generated_trajectories}

\begin{figure}[!t]
    \centering
    \includegraphics[width=0.9\linewidth]{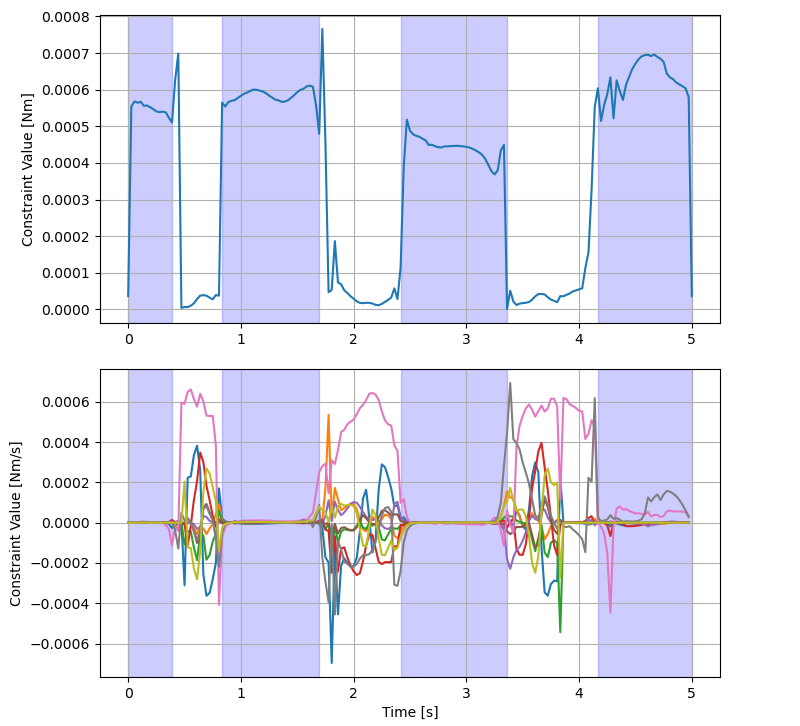}
    \vspace{-2mm}
    \caption{Time evolution of the contact condition (up) and the nine components of no-slip condition (down) for a single contact point during walking on undulating terrain (\Cref{fig:trajectories}(c)), with relaxation intervals of $[0,\, 0.001]$~\si{\newton\meter} and $[-0.001,\, +0.001]$~\si{\newton\meter\per\second} respectively. Shaded areas indicate contact phases. The point is considered to be in contact when its vertical distance from the ground is below \SI{1}{\milli \meter}.}
    \label{fig:quantitative_analysis}
\end{figure}

\begin{figure*}[!t]
    \centering
    \begin{minipage}{0.9\textwidth}
        \begin{overpic}[width=\linewidth]{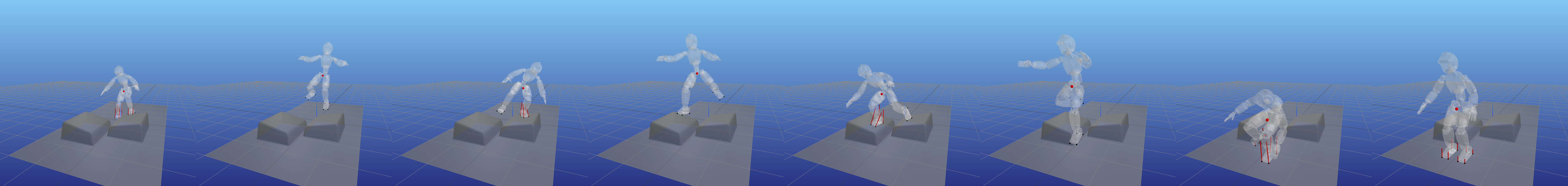}
            \put(0.5,1){\textcolor{white}{(a)}}
        \end{overpic}
    \end{minipage}
    \vspace{5pt}
    
    \begin{minipage}{0.9\textwidth}
        \begin{overpic}[width=\linewidth]{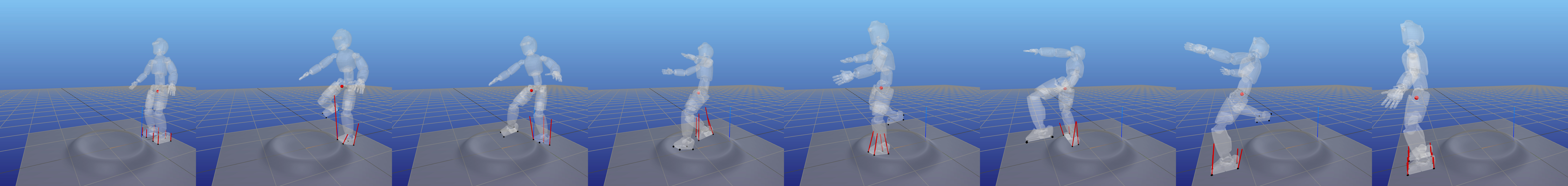}
            \put(0.5,1){\textcolor{white}{(b)}}
        \end{overpic}
    \end{minipage}
    \vspace{5pt}
    
    \begin{minipage}{0.9\textwidth}
        \begin{overpic}[width=\linewidth]{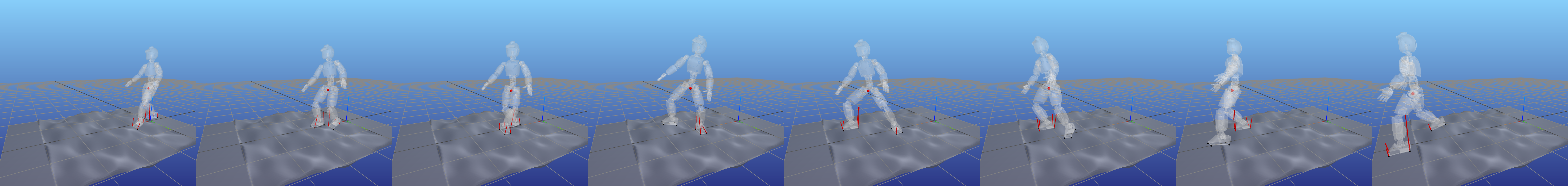}
            \put(0.5,1){\textcolor{white}{(c)}}
        \end{overpic}
    \end{minipage}
    \vspace{5pt}
    
    \begin{minipage}{0.9\textwidth}
        \begin{overpic}[width=\linewidth]{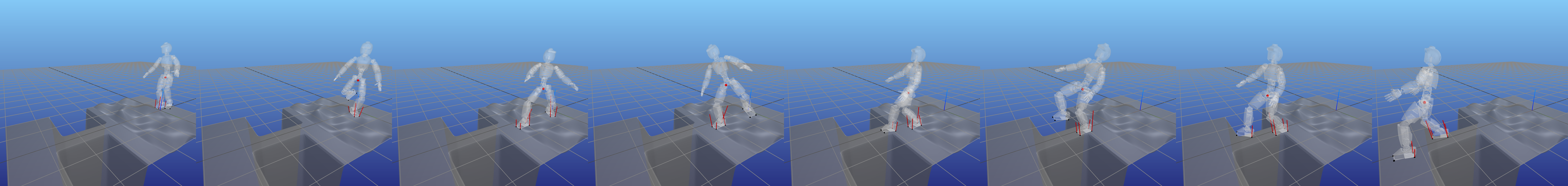}
            \put(0.5,1){\textcolor{white}{(d)}}
        \end{overpic}
    \end{minipage}
    \vspace{5pt}
    
    \caption{Examples of generated trajectories: (a) inclined platform jumps, (b) crater walk, (c) undulating terrain walk, and (d) challenging terrain walk.}
    \label{fig:trajectories}
\end{figure*}

\begin{itemize}[align=right,itemindent=2em,leftmargin=0pt]
\item \emph{Stepping Up/Down a Step:} The robot steps up/down a $\SI{20}{\centi\meter}$ high step. During this movement, the CoM vertical projection stays outside the support polygon, indicating a dynamic rather than quasi-static gait. The Swing Height Task ensures the robot lifts its leg enough to avoid unexpected collisions with the step edge.
\item \emph{Walking on Complex Terrain:}
The robot successfully walks on ramps and complex terrain using the same initial guess as for flat-ground walking, highlighting the planner's versatility. Only the terrain information needs to be changed. Task \eqref{eq:lateral_distance_task} prevents the legs from self-penetrating. Since the surface is curved, partial foothold contacts and contact on differently oriented planar surfaces are common (for instance, in the 7\textsuperscript{th} snapshot of \Cref{fig:trajectories}(a)). Penetration may occur in concave areas, as the planner only enforces that the four vertices of the foot do not penetrate the ground. This issue can be resolved by adding constraints to ensure that other internal points of the sole do not penetrate the terrain.
\item \emph{Jumping:}
The robot performs jumps on flat terrain or on jumping platforms. As noted earlier, the initial guess is crucial and must feature zero forces during the flight phase. During the take-off and landing phases, the contact forces of the initial guess are scaled by a factor of 2 to 15 times the static equilibrium value to induce a net CoM acceleration. For the front flip, joint regularization was adjusted to promote a crouched posture during the aerial phase.
\end{itemize}
Four samples of generated trajectories are shown in \Cref{fig:trajectories}, with related data reported in \Cref{table:example_data}. Additional data and a wider range of dynamic motions appear in the supplementary video. \Cref{fig:quantitative_analysis} shows the time evolution of the contact and no-slip conditions for a single foot vertex during walking on undulating terrain (\Cref{fig:trajectories}(c)). The relaxation intervals were set to $[0,\, 0.001]$~\si{\newton\meter} and $[-0.001,\, +0.001]$~\si{\newton\meter\per\second} respectively. These constraints are never violated during the motion. 

Compared with the method presented by Winkler \emph{et al.} \cite{winkler2018gait}, our approach offers greater flexibility by allowing the number of contact events to be determined within the optimization rather than fixed a priori. By explicitly employing a smooth mathematical model of the terrain integrated with CCs, the planner can manage and exploit an arbitrary number of contact interactions that are not predetermined, which is crucial for planning highly dynamic motions on uneven surfaces. This expressiveness, however, comes at the cost of increased computational complexity and potential numerical challenges.

\begin{table}[!t]
    \centering
    \caption{Trajectory generation: example data (\Cref{fig:trajectories}).}
    \label{table:example_data}
    \renewcommand{\arraystretch}{1.2}
    \rowcolors{1}{}{gray!10}
    \begin{tabular}{lccccccc}
        \toprule
        & \textbf{T} [\si{\second}] & \textbf{N} & \textbf{MTM} & \textbf{CIP} & \textbf{CV} & \textbf{CT} [\si{\second}] & \textbf{I}\\
        \midrule
        \textbf{(a)} & $5.0$ & $60$ & & & $\SI{1.13e{-4}}{}$ & $75.8$ & $350$ \\
        \textbf{(b)} & $6.0$ & $60$ & \checkmark & \checkmark & $\SI{3.57e{-4}}{}$ & $55.3$ & $250$ \\
        \textbf{(c)} & $5.0$ & $180$ & \checkmark & \checkmark & $\SI{1.42e{-3}}{}$ & $186.6$ & $250$ \\
        \textbf{(d)} & $9.0$ & $180$ & \checkmark & \checkmark & $\SI{4.26e{-3}}{}$ & $173.3$ & $250$ \\
        \bottomrule
    \end{tabular}
    \\[5pt]
    \footnotesize{T: \textit{time horizon}; N: \textit{number of time intervals}; MTM: \textit{use of the motor torque minimization task}; CIP: \textit{use of the constraint on the sole internal points}; CV: \textit{solver constraint violation}; CT: \textit{solver computation time; I: solver iterations.}}
\end{table}

\subsection{Tested Trajectories}
\label{sec:results:tested_trajectories}
Low-dynamic motions have been tested both in simulation and on the humanoid robot ErgoCub (\Cref{fig:ergoCub}). The trajectories, generated offline, are provided as a reference to a joint position PID controller. Thus, loop closure occurs exclusively at the joint level, with no guarantee of correct tracking of the floating-base pose. Since the sampled trajectory frequency is around $\SI{10}{\hertz}$, we increased the sample density by a factor of $10$-$20$ through linear interpolation. The walking and step-climbing (height: $\SI{7}{\centi \meter}$) motions have been proven to be remarkably accurate and have been successfully reproduced in the simulation and on the real robot, even in the absence of a high-level closed-loop controller. The robot performed the tasks without losing balance, thus validating the physical feasibility of the trajectories generated by the planner. The optimization yields feasible trajectories that naturally involve edge contacts, without requiring explicit enforcement. In this case, heel contact precedes toe contact, resulting in a gait pattern more closely resembling that of humans. Some trajectories involving contacts with a ramp have been successfully reproduced in simulation. More complex trajectories cannot be adequately executed as long as an open-loop control architecture is used. The supplementary video shows experiments conducted both in simulation and in the real world.

\section{Conclusions and Future Work}
\label{sec:conclusions}
The paper presents an offline whole-body trajectory planner for humanoid locomotion on uneven terrain, framed as a constrained OC problem solved via direct transcription. The terrain was modeled using an originally devised representation for smooth surfaces, generalizing contact constraints beyond flat surfaces. The generated trajectories demonstrate the effectiveness of this approach, which enables dynamic motions on complex terrain. Future work will extend the real-robot validation to more dynamic motions once a high-level closed-loop controller is in place to ensure accurate tracking. Another possibility is to integrate the planner with RL modules. For example, the generated trajectories could serve as valuable motion priors for training learning-based controllers.

Regarding the limitations of the proposed model, it remains effective as long as real terrains can be approximated by relatively simple, easily parameterizable shapes. Additionally, its reliance on explicit functions for geometry description restricts the range of representable walking surfaces. Employing implicit functions could enable a richer set of geometric shapes. Nevertheless, while the current approach remains highly effective in structured, human-designed environments, developing a method to generate differentiable terrains from point clouds would enable planning over arbitrary, unstructured terrains unknown a priori. Finally, the current foot contact model (four contact points per foot) and lack of self-collision checks limit accuracy and realism.

\bibliographystyle{IEEEtran}
\bibliography{IEEEabrv, bibliography}
\end{document}